\documentclass{article}

\usepackage{microtype}
\usepackage{graphicx}
\usepackage{subcaption}
\usepackage{booktabs} % for professional tables

\usepackage{hyperref}
\usepackage{enumitem}

\usepackage[preprint]{icml2026}

\usepackage{amsmath}
\usepackage{amssymb}
\usepackage{mathtools}
\usepackage{amsthm}
\usepackage{multirow}
\usepackage{multicol}

\usepackage{enumitem}
\usepackage{color}
\usepackage{xcolor}  
\usepackage{colortbl}

\usepackage[capitalize,noabbrev]{cleveref}

\theoremstyle{plain}
\newtheorem{theorem}{Theorem}[section]
\newtheorem{proposition}[theorem]{Proposition}

\theoremstyle{definition}

\theoremstyle{remark}

\usepackage[textsize=tiny]{todonotes}

\icmltitlerunning{On the Plasticity and Stability for Post-Training Large Language Models}

\begin{document}

\twocolumn[
  \icmltitle{On the Plasticity and Stability for Post-Training Large Language Models}

  \icmlsetsymbol{corr}{*}

  \begin{icmlauthorlist}
    \icmlauthor{Wenwen Qiang}{1,2}
    \icmlauthor{Ziyin Gu}{1,2}
    \icmlauthor{Jiahuan Zhou}{3}
    \icmlauthor{Jie Hu}{4}
    \icmlauthor{Jingyao Wang}{1,2,corr}
    \icmlauthor{Changwen Zheng}{1,2}
    \icmlauthor{Hui Xiong}{5}
  \end{icmlauthorlist}

\icmlaffiliation{1}{Institute of Software Chinese Academy of Sciences, Beijing, China}
\icmlaffiliation{2}{University of the Chinese Academy of Sciences, Beijing, China}
\icmlaffiliation{3}{Wangxuan Institute of Computer Technology, Peking University, Beijing, China}
\icmlaffiliation{4}{Meituan}
\icmlaffiliation{5}{Thrust of Artificial Intelligence, The Hong Kong University of Science and Technology (Guangzhou), China
Department of Computer Science and Engineering, The Hong Kong University of Science and Technology Hong Kong SAR, China}

  \icmlcorrespondingauthor{Jingyao Wang}{wangjingyao2023@iscas.ac.cn}

  \vskip 0.3in
]

% this must go after the closing bracket ] following \twocolumn[ ...

% This command actually creates the footnote in the first column listing the
% affiliations and the copyright notice. The command takes one argument, which
% is text to display at the start of the footnote. The \icmlEqualContribution
% command is standard text for equal contribution. Remove it (just {}) if you
% do not need this facility.

% Use ONE of the following lines. DO NOT remove the command.
% If you have no special notice, KEEP empty braces:
\printAffiliationsAndNotice{}  % no special notice (required even if empty)
% Or, if applicable, use the standard equal contribution text:
% \printAffiliationsAndNotice{\icmlEqualContribution}

\begin{abstract}
Training stability remains a critical bottleneck for Group Relative Policy Optimization (GRPO), often manifesting as a trade-off between reasoning plasticity and general capability retention. We identify a root cause as the geometric conflict between plasticity and stability gradients, which leads to destructive interference. Crucially, we argue that deterministic projection methods are suboptimal for GRPO as they overlook the intrinsic stochasticity of group-based gradient estimates. To address this, we propose \textbf{P}robabilistic \textbf{C}onflict \textbf{R}esolution (\textbf{PCR}), a Bayesian framework that models gradients as random variables. PCR dynamically arbitrates conflicts via an uncertainty-aware ``soft projection'' mechanism, optimizing the signal-to-noise ratio. Extensive experiments demonstrate that PCR significantly smooths the training trajectory and achieves a superior performance in various reasoning tasks.
\end{abstract}

\section{Introduction}
\label{sec:intro}

Large Language Models (LLMs) like DeepSeek-R1~\citep{guo2025deepseek} has proven that Reinforcement Learning (RL) is essential for unlocking complex reasoning capabilities. Among the various RL techniques, Group Relative Policy Optimization (GRPO)~\citep{shao2024deepseekmath} has become the standard choice. By removing the need for a separate value network, GRPO significantly reduces memory usage and enables scalable training, allowing models to learn effectively from group-based relative rewards.

However, despite its efficiency, GRPO is notoriously difficult to train~\cite{zhang2025gvpo,ge2025grpo,wu2025takes}. Practitioners face a harsh ``Plasticity-Stability Dilemma'': aggressive updates to improve reasoning (plasticity) often cause the model to forget general knowledge or lose linguistic coherence (stability). Conversely, strict constraints to preserve language often prevent the model from learning new reasoning skills. Balancing these two forces typically requires exhaustive, fragile tuning of the KL penalty coefficient ($\beta$).

Why is GRPO so unstable? In this paper, inspired by \cite{liu2021conflict}, we argue that a possible root cause is a geometric conflict between two optimization objectives in the GRPO loss. Our analysis reveals that the gradient for reasoning ($\mathbf{g}_{pla}$) and the gradient for stability ($\mathbf{g}_{sta}$) frequently point in opposite directions. The standard GRPO update rule simply adds these two opposing vectors together. This results in ``destructive interference'', where the two forces cancel each other out, causing the optimizer to fight against itself and traverse the loss landscape inefficiently.

Addressing this conflict is not as simple as using geometric projection methods like PCGrad~\citep{yu2020gradient}. These methods assume that the calculated gradients are perfect, deterministic vectors. However, in GRPO, gradients are Monte Carlo estimates derived from a small group of training queries, Therefore, they are inherently noisy and uncertain. If we blindly project one gradient vector onto another based on noisy data (a ``hard projection''), we risk discarding valid learning signals or enforcing incorrect constraints. We need a method that considers not just the direction of the gradients, but our confidence in them.

To solve this, we propose \textbf{P}robabilistic \textbf{C}onflict \textbf{R}esolution (\textbf{PCR}). Instead of treating gradients as fixed arrows, we model them as probability distributions (Gaussian random variables) to capture their uncertainty. We then employ Bayesian inference to dynamically arbitrate the conflict. The core idea is intuitive: if the reasoning signal is strong and precise (low variance), PCR trusts it and allows the update; if the reasoning signal is noisy or the stability constraint is rigid (high variance), PCR suppresses the update. This results in a soft projection mechanism. Unlike PCGrad, which deletes conflicting components entirely, PCR scales them based on the signal-to-noise ratio. To make this computationally feasible for billion-parameter models, we apply PCR strategically: we use it only on the MLP layers (which act as knowledge stores) while using standard updates for Attention layers. This hybrid approach ensures rigorous stability for core knowledge without slowing down training.

Our contributions: \textbf{1}) We identify that the instability in GRPO stems from the high-dimensional geometric conflict between plasticity and stability gradients; \textbf{2}) We propose PCR, a Bayesian framework that derives a closed-form, uncertainty-aware ``soft projection'' rule. To our knowledge, this is the first work to introduce probabilistic modeling into gradient projection for LLM post-training; \textbf{3}) We introduce an efficient hybrid implementation that applies PCR only to MLP layers, making it scalable for LLM training; \textbf{4}) Theoretically, we prove that PCR is the mathematically optimal estimator in the gradient space. It minimizes the update error by finding the perfect trade-off between bias and variance. Empirically, extensive experiments demonstrate that this stability allows PCR to eliminate training oscillations and achieve superior performance on reasoning tasks.

\section{Related Work}
\label{sec:related_work}

Large language models (LLMs) have achieved remarkable success in reasoning tasks through reinforcement learning post-training~\cite{ouyang2022training,wang2025learning}. While GRPO~\cite{shao2024deepseekmath} has emerged as a dominant paradigm due to its memory efficiency, training stability remains a critical bottleneck. 
Recent studies have proposed various mechanisms to mitigate this volatility. 
On one hand, methods like $\Delta$L Norm~\cite{he2025deltal} and BNPO~\cite{xiao2025bnpo} introduce advanced normalization statistics to smooth the reward landscape and reduce variance. 
On the other hand, approaches like GSPO~\cite{zheng2025group} and BAPO~\cite{xi2025bapo} focus on refining policy constraints, employing adaptive clipping or sequence-level optimization to prevent policy divergence. 
Other works such as GVPO~\cite{zhang2025gvpo} and MRT~\cite{qu2025optimizing} utilize analytic re-weighting or reward correction. Meanwhile, GCPO~\cite{,gu2025group} maintains stable training by constructing causal collision structures.
Despite these diverse improvements, most existing methods focus on shaping the scalar loss or reward values. They often overlook the high-dimensional geometric antagonism between the plasticity gradient and the stability constraint, which our work identifies as a possible root cause of optimization conflict.

The challenge of conflicting gradients is a central theme in Multi-Task Learning (MTL). When objectives compete, simple summation leads to destructive interference. 
Classic solutions like PCGrad~\cite{yu2020gradient} project conflicting gradients onto the normal plane, while recent advances like MMPareto~\cite{wei2024mmpareto} and Robust MTL~\cite{he2024robust} explore Pareto-optimal frontiers and risk minimization to balance innocent unimodal assistance or excess risks.
However, these methods typically operate under a deterministic assumption, treating gradient estimates as reliable ground truths. In the context of GRPO, gradients are Monte Carlo estimates carrying significant stochastic noise. Applying hard geometric projection to such noisy signals can erroneously discard valid exploration. 
Our proposed PCR bridges this gap by modeling gradients as probabilistic distributions. Instead of a hard cut, PCR introduces a Bayesian arbitration mechanism that performs ``soft projection'' based on the signal-to-noise ratio, offering a mathematically optimal trade-off for stochastic optimization.

\section{Problem Formulation and Analysis}
\label{sec:preliminaries}

To rigorously analyze the plasticity-stability trade-off in post-training, we first reformulate the GRPO objective by decoupling it into two distinct forces. We then derive their respective gradients to examine their physical roles. Finally, through empirical analysis, we demonstrate that the inherent geometric conflict between these update directions serves as the root cause of training instability. 

\subsection{Reformulating the GRPO Objective}

A LLM can be formulated as an autoregressive policy $\pi_\theta(\cdot\mid q)$, which generates a response token-by-token conditioned on a query $q$. GRPO optimizes this policy by introducing a group relative advantage. Given a query $q \sim P$ and a group of candidate outputs $\{y_i\}_{i=1}^{n}$ sampled from the old policy $\pi_{\theta_{\rm old}}$, the standard GRPO objective aims to maximize:
\begin{equation}
\label{eq_grpo_obj}
\begin{split}
\mathcal{J}_{\rm GRPO}(\theta) = \mathbb{E}_{q \sim P, \{y_i\} \sim \pi_{\theta_{\rm old}}} [ \frac{1}{n} {\textstyle \sum_{i=1}^{n}} \frac{1}{T_i} \\
{\textstyle \sum_{j=1}^{T_i}} \left( \mathcal{S}_{i,j}(\theta) - \beta \mathcal{K}_{i,j}(\theta) \right) ].
\end{split}
\end{equation}
To clearly identify the sources of the gradients, we formally define the two core components in the objective above:

(1) \textbf{Token-level Surrogate Gain ($\mathcal{S}_{i,j}$):} This term drives the model updates towards regions of higher reward. We define the importance ratio as $R_{i,j}(\theta) = \frac{\pi_{\theta}(y_{i,j}\mid q,y_{i,<j})}{\pi_{\theta_{\rm old}}(y_{i,j}\mid q,y_{i,<j})}$. Incorporating the clipping mechanism for training stability, the surrogate gain is defined as:
\begin{equation}
\resizebox{0.9\linewidth}{!}{$
\mathcal{S}_{i,j}(\theta) = \min \left( R_{i,j}(\theta) A_i, \; \text{clip}(R_{i,j}(\theta), 1-\epsilon, 1+\epsilon) A_i \right),
$}
\end{equation}
where $A_i$ denotes the group relative advantage, computed by standardizing the rewards within the group: $A_i = (r_i - \mu_{\text{group}}) / \sigma_{\text{group}}$, and $y_i=(y_{i,1},\cdots,y_{i,T_i})$.

(2) \textbf{Token-level KL Penalty ($\mathcal{K}_{i,j}$):} This term acts as a regularizer, constraining the policy from deviating excessively from the reference distribution:
\begin{equation}
\mathcal{K}_{i,j}(\theta) = D_{\rm KL} \left( \pi_\theta(\cdot|q, y_{i,<j}) \parallel \pi_{\rm ref}(\cdot|q, y_{i,<j}) \right).
\end{equation}
Here, $\pi_{\rm ref}$ is typically set to the old policy $\pi_{\theta_{\rm old}}$.

\subsection{Dual Decomposition of Loss and Gradients}

For gradient analysis, we transform the maximization problem of $\mathcal{J}_{\rm GRPO}$ into a minimization problem of a loss function $\mathcal{L}_{\rm GRPO} = -\mathcal{J}_{\rm GRPO}$. Based on the Eq. \ref{eq_grpo_obj}, we explicitly decompose the total loss into two independent terms representing plasticity and stability:
\begin{equation}
\mathcal{L}_{\rm GRPO}(\theta) = \underbrace{\mathcal{L}_{\rm pla}(\theta)}_{\text{Plasticity}} + \beta \cdot \underbrace{\mathcal{L}_{\rm sta}(\theta)}_{\text{Stability}},
\end{equation}
Here, the plasticity loss corresponds to the negative expectation of the surrogate gain. It aims to enhance task-specific performance by exploiting the advantage signals:
    \begin{equation}
    \mathcal{L}_{\rm pla}(\theta) = - \mathbb{E} [ \frac{1}{n} {\textstyle \sum_{i=1}^{n}} \frac{1}{T_i} {\textstyle \sum_{j=1}^{T_i}} \mathcal{S}_{i,j}(\theta) ].
    \end{equation}
The stability loss corresponds to the KL divergence term. It aims to anchor the policy to the reference manifold to preserve general capabilities:
    \begin{equation}
    \mathcal{L}_{\rm sta}(\theta) = \mathbb{E} [ \frac{1}{n} {\textstyle \sum_{i=1}^{n}} \frac{1}{T_i} {\textstyle \sum_{j=1}^{T_i}} \mathcal{K}_{i,j}(\theta) ].
    \end{equation}

From the above, we derive two gradient vectors used for parameter updates. Relying on the linearity of the gradient operator, the total gradient $\mathbf{g}_{total}$ can be expressed as:
\begin{equation}
\nabla_\theta \mathcal{L}_{\rm GRPO}(\theta) = \mathbf{g}_{pla} + \beta \cdot \mathbf{g}_{sta}.
\end{equation}
Suppose a batch has $N_{\text{batch}}$ queries, these two gradients possess distinct physical interpretations and adversarial natures:

(1) \textbf{The Plasticity Gradient ($\mathbf{g}_{pla}$):}
\begin{equation}
\mathbf{g}_{pla} \triangleq \nabla_\theta \mathcal{L}_{\rm pla}(\theta) \approx - \frac{1}{N_{\text{batch}}} \sum \nabla_\theta \mathcal{S}_{i,j}(\theta).
\end{equation}
This gradient represents the signal for policy improvement. It drives the parameters $\theta$ in a direction that maximizes the specific task reward, serving as the primary source of the model's new capabilities (i.e., plasticity).

(2) \textbf{The Stability Gradient ($\mathbf{g}_{sta}$):}
\begin{equation}
\mathbf{g}_{sta} \triangleq \nabla_\theta \mathcal{L}_{\rm sta}(\theta) \approx \frac{1}{N_{\text{batch}}} \sum \nabla_\theta \mathcal{K}_{i,j}(\theta).
\end{equation}
This gradient represents the signal for behavior maintenance. It drives the parameters $\theta$ back towards the parameter space of the reference model, serving as the key constraint to prevent catastrophic forgetting and maintain general capabilities (i.e., stability).

\subsection{Empirical Analysis}
\label{sec:empirical_findings}

While GRPO theoretically balances policy improvement and reference maintenance, in practice, it exhibits significant training instability \cite{ouyang2022training,simoni2025gtpo,dai2025stable}. Models often oscillate between overfitting to the reasoning task (forgetting general knowledge) and stagnation (failing to learn reasoning), requiring exhaustive hyperparameter tuning. To diagnose the root cause of this instability, we analyze the optimization dynamics of DeepSeek-R1-Distill-Llama-8B on the AIME dataset (detailed setup in Appendix B). 

\begin{figure*}
    \centering

    \begin{subfigure}[b]{0.32\textwidth}
    \centering
    \includegraphics[width=\textwidth]{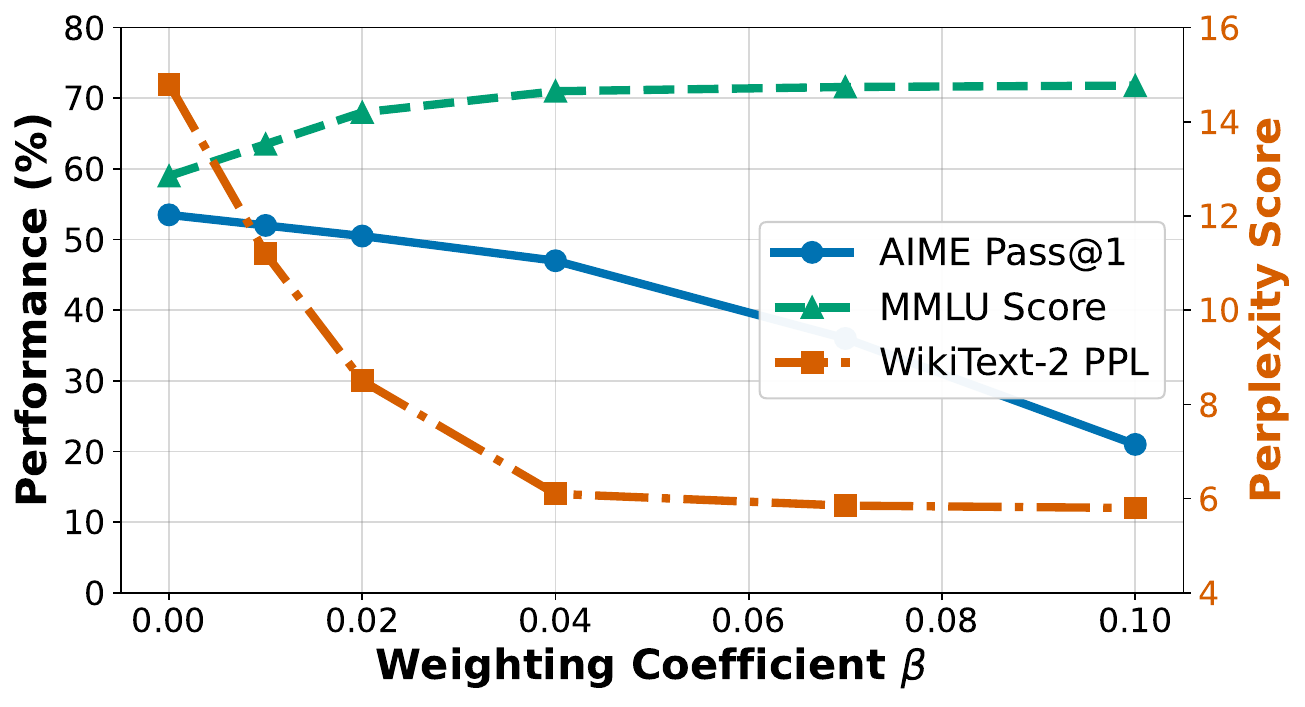}
    % \label{fig:gradient_roles}
    \caption{Trade-off performance.}
  \end{subfigure}
  \hfill
  \begin{subfigure}[b]{0.32\textwidth}
    \centering
    \includegraphics[width=\textwidth]{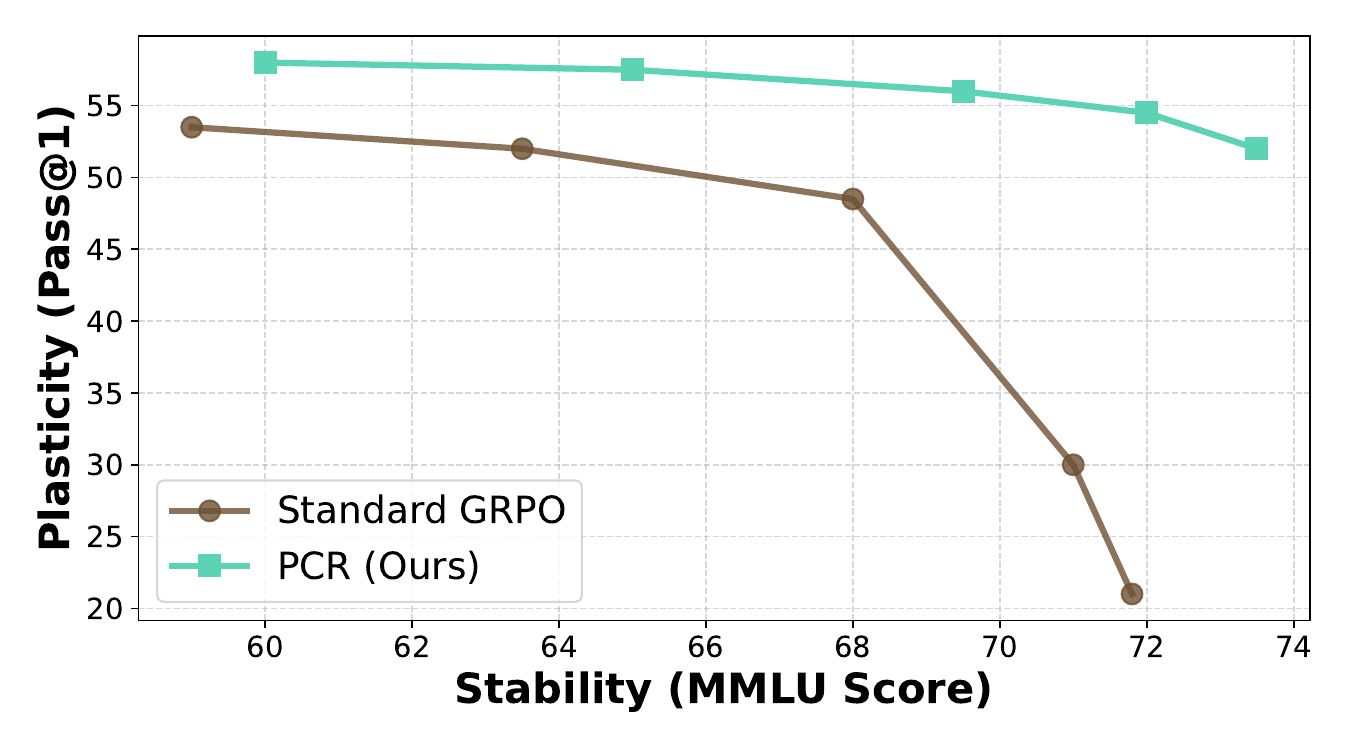}
    \caption{The Pareto frontier.}
  \end{subfigure}
  \hfill
  \begin{subfigure}[b]{0.32\textwidth}
    \centering
    \includegraphics[width=\textwidth]{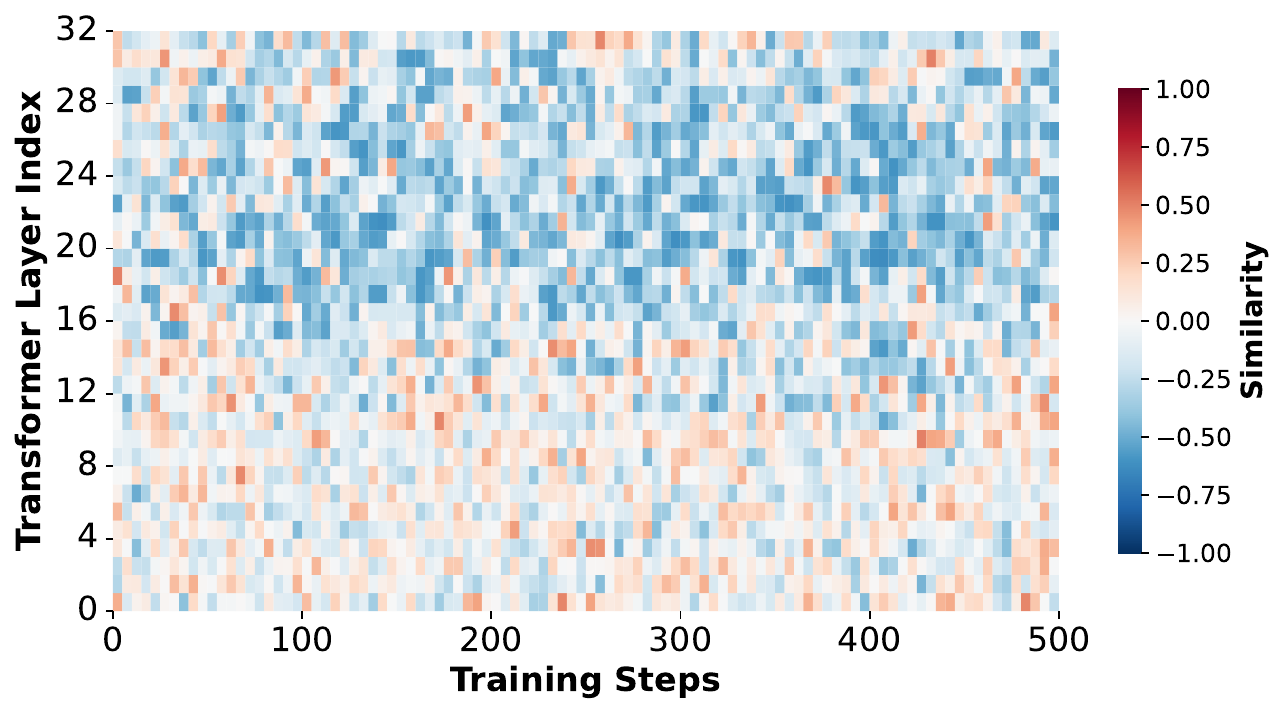}
    \caption{Visualization of heatmap.}
  \end{subfigure}

    \caption{Motivating results. (a) Results of AIME accuracy, MMLU score, and PPL, varying with the KL coefficient $\beta$. (b) The Pareto frontier. (c) The layer-wise cosine similarity between plasticity and stability gradients across training steps.}
    \label{fig:motivation_experiments}
    \vspace{-0.15 cm}
\end{figure*}

We first visualize the trade-off between reasoning performance (AIME Pass@1) and linguistic stability (WikiText-2 PPL) by varying the KL coefficient $\beta$. 
The results are shown in Figure~\ref{fig:motivation_experiments}(a)-(b).
We can observe that a slight decrease in $\beta$ triggers a disproportionate collapse in PPL (Note that a lower PPL is better), while a slight increase suppresses reasoning gains entirely. 
As shown in Figure~\ref{fig:motivation_experiments}(b), the Pareto frontier is extremely sharp.
This sensitivity suggests that the two objectives are functionally antagonistic, and standard scalarization fails to find a stable equilibrium, leading to the observed training volatility.

To confirm this antagonism physically, we compute the layer-wise cosine similarity between the Plasticity gradient $\mathbf{g}_{pla}$ and the stability gradient $\mathbf{g}_{sta}$. The heatmap in Figure~\ref{fig:motivation_experiments}(b) reveals that the middle-to-deep MLP layers—critical for knowledge storage—exhibit persistent negative cosine similarity (gradient conflict) throughout training.
This geometric conflict implies that the standard GRPO update rule ($\mathbf{g}_{pla} + \beta \mathbf{g}_{sta}$) results in destructive interference: the opposing vectors partially cancel each other out, reducing the effective update magnitude and skewing the direction. This cancellation explains the instability: the optimizer is fighting against itself, leading to inefficient traversal of the loss landscape and sensitivity to noise.

\section{Methodology}
\label{sec:method}

In this section, we introduce Probabilistic Conflict Resolution (PCR), a principled framework that dynamically arbitrates gradient conflicts in GRPO. Our method unfolds in four key stages: probabilistic modeling, geometric decomposition, Bayesian arbitration, and finally reconstruction and projection. This pipeline effectively transforms the hard geometric constraints of prior methods into a flexible, uncertainty-aware optimization process.

\subsection{Probabilistic Modeling}

As formalized in the previous section, the plasticity gradient $\mathbf{g}_{pla}$ and the stability gradient $\mathbf{g}_{sta}$ frequently exhibit adversarial directions. The canonical approach to resolving such conflicts is PCGrad~\cite{yu2020gradient}, which treats gradients as deterministic vectors and performs geometric projection upon conflict. However, this deterministic perspective overlooks a fundamental property of GRPO optimization: the observed gradients are not true expectations over the full data distribution but are Monte Carlo estimators derived from a finite group of samples.

From a statistical lens, gradient estimates are often noisy. A gradient with high variance is unreliable, meaning we cannot fully trust its direction. In such cases, PCGrad blindly applies projection regardless of uncertainty, creating the risk of over-trusting a noisy direction and removing valid signals. Intuitively, the projection should be adaptive based on confidence. If the gradient is precise (low variance), we should strictly perform geometric projection. However, if the gradient is noisy and random (high variance), we should trust it less and perform a weaker projection to avoid being misled by noise. Therefore, to formalize this adaptive mechanism, we must model gradients as random variables rather than deterministic vectors, which allows us to quantify the uncertainty (refer to Appendix C for more explanation). 

To mathematically capture this uncertainty, we look at the aggregation mechanism of GRPO. Since the gradient is computed by averaging over a group of independent stochastic queries $\{q_i\}_{i=1}^{N_{\rm batch}}$, we can invoke the Central Limit Theorem (CLT)~\citep{feller1991introduction}. The CLT states that the distribution of the sample mean approaches a multivariate Gaussian as the sample size implies. This provides a rigorous justification for approximating the gradient estimation as Gaussian, a standard practice in the analysis of stochastic optimization dynamics~\citep{mandt2017stochastic}. By leveraging this property, we can explicitly quantify the confidence of update directions via their covariance structure. Thus, we have:
\begin{equation}
\resizebox{0.89\linewidth}{!}{$
\mathbf{g}_{pla} \sim \mathcal{N}(\boldsymbol{\mu}_{pla}, \Sigma_{pla}), \; \mathbf{g}_{sta} \sim \mathcal{N}(\boldsymbol{\mu}_{sta}, \Sigma_{sta}),
$}
\end{equation}
where $\boldsymbol{\mu}_{pla}$ and $\boldsymbol{\mu}_{sta}$ represent the theoretical expectations of the gradients over the underlying data distribution, e.g., the latent ground truths we aim to approximate. Meanwhile, $\Sigma$ characterize the estimation uncertainty, quantifying how significantly the observed gradients derived from group sampling diverge from these true expectations.

To bridge the gap between theoretical rigor and practical feasibility, we adopt three physically motivated approximations. First, regarding the direction, we approximate the latent true mean $\boldsymbol{\mu}$ directly using the observed gradient $\mathbf{g}$. This relies on the fundamental premise of stochastic approximation~\cite{robbins1951stochastic}, treating the empirical gradient as an unbiased estimator of the population expectation. Second, following standard adaptive optimization practices (e.g., Adam~\cite{kingma2014adam}), we simplify the complex covariance structure via an isotropic assumption ($\Sigma \approx \sigma^2 \mathbf{I}$), where the scalar variance is estimated by the trace of intra-group gradients. This effectively captures the global noise level without incurring memory overhead. Third, inspired by the mean-field assumption common in variational inference~\cite{blei2017variational}, we treat the estimation errors of the two gradients as conditionally independent. This is justified by their orthogonal sources of stochasticity: the variance in $\mathbf{g}_{pla}$ is primarily driven by the instability of discrete rewards, whereas the variance in $\mathbf{g}_{sta}$ arises from the probabilistic divergence of the token distribution.

\subsection{Geometric Decomposition}

While the probabilistic model measures gradient uncertainty, the actual interaction between the two gradient distribution is determined by their expected directions, represented by the means $\boldsymbol{\mu}_{pla}$ and $\boldsymbol{\mu}_{sta}$ (refer to Appendix E for more detailed explanation). To rigorously incorporate uncertainty into conflict resolution, we must first analyze the geometric relationship between these two vectors. The goal of this step is to physically isolate the specific component responsible for the directional opposition.

We formally define a gradient conflict as a situation where the two expected gradients point in opposing directions, satisfying $\boldsymbol{\mu}_{pla} \cdot \boldsymbol{\mu}_{sta} < 0$.  This simply means that the plasticity objective is pushing the parameters in a direction that hurts the stability objective. To resolve this, mere identification is not enough. We must geometrically separate the ``safe'' part of the update from the ``harmful'' part.

To isolate the specific source of the conflict, we perform an orthogonal decomposition of the plasticity gradient $\boldsymbol{\mu}_{pla}$ using the stability gradient $\boldsymbol{\mu}_{sta}$ as the reference frame. We choose the stability gradient as the anchor because of the Alignment Hypothesis~\citep{zhou2024lima}. It suggests that the pre-trained knowledge is the foundation of the model's intelligence. Therefore, our primary goal is to learn new capabilities without breaking this foundation. Based on this logic, we decompose the plasticity gradient as follows:
\begin{equation}
\boldsymbol{\mu}_{pla} = \boldsymbol{\mu}_{pla}^{\perp} + \boldsymbol{\mu}_{pla}^{\parallel}.
\end{equation}
This formula splits the update into two components with clear physical meanings. The first is the independent component ($\boldsymbol{\mu}_{pla}^{\perp}$), which is perpendicular to the stability gradient. Since it moves in a direction that has no effect on the stability objective, it is structurally safe and should be fully preserved. The second is the conflicting component ($\boldsymbol{\mu}_{pla}^{\parallel}$), which is the projection of the plasticity gradient onto the stability gradient. When a conflict happens, this component points directly against the constraint:
\begin{equation}
\boldsymbol{\mu}_{pla}^{\parallel} = \frac{\boldsymbol{\mu}_{pla} \cdot \boldsymbol{\mu}_{sta}}{\|\boldsymbol{\mu}_{sta}\|^2} \boldsymbol{\mu}_{sta}.
\end{equation}
This component represents the specific force that attempts to violate the stability rule. Note that in this step, we only identify this conflicting force geometrically. The decision of whether to remove it or keep it depends on its uncertainty, which we will handle in the next section.

\subsection{Bayesian Arbitration}

Having isolated the conflicting component $\boldsymbol{\mu}_{pla}^{\parallel}$, we face the central decision of determining exactly how much of this conflicting update to retain. Traditional methods like PCGrad are too extreme because they bluntly discard the entire component, implicitly assuming that the stability constraint is perfect and allows for no error. To correct this bias, we use Bayesian inference to dynamically balance the two objectives based on their reliability. Crucially, although the parameter space of an LLM is high-dimensional, the gradient conflict is strictly localized along a single line defined by the stability gradient $\boldsymbol{\mu}_{sta}$. Any movement perpendicular to this axis is already safe. Therefore, we can simplify this complex optimization into a scalar estimation problem for a variable $x$, denoting the optimal move along conflict axis.

To estimate this optimal magnitude, we combine two distinct sources of information. We view the plasticity gradient as a noisy observation (the Likelihood). It pushes the model to move by a magnitude of $x_{obs} = \|\boldsymbol{\mu}_{pla}^{\parallel}\|$, but its reliability is inherently limited by its variance. If this driving signal is precise, we trust the observation and move. Simultaneously, we view the stability requirement as a prior belief (the Prior). Its goal is to act as an anchor, aiming to prevent deviation from the reference manifold. Mathematically, this creates a preference for the update to be zero. The variance of this prior acts like a physical stiffness parameter. A small variance represents a rigid wall that forbids movement, while a large variance acts like a flexible elastic band that permits necessary exploration.

By combining the likelihood and the prior using Bayes' theorem, the optimal update $x^*$ emerges as the precision-weighted average of the two. We formalize this result as:

\begin{proposition}[Optimal Conflict Retention]
The optimal update magnitude $x^*$ is governed by the following:
\begin{equation}
x^* = k \cdot x_{obs}, \quad \text{where} \quad k = \frac{\lambda_{pla}}{\lambda_{pla} + \lambda_{sta}}.
\end{equation}
Here, $\lambda = 1/\sigma^2$ denotes precision, and the scalar $k \in [0, 1]$ is defined as the retention coefficient.
\end{proposition}

The proof is provided in Appendix F. The coefficient $k$ offers a very clear physical interpretation: it represents the relative confidence of the plasticity signal compared to the stability constraint. If the plasticity signal is far more reliable, $k$ approaches 1 and we fully keep the update. Conversely, if the stability constraint is far more certain, $k$ approaches 0 and we discard the update. This allows the algorithm to smoothly interpolate between exploring and restraining based on the instantaneous signal-to-noise ratio.

\subsection{Reconstruction and Projection}

With the retention coefficient $k$ derived from Bayesian arbitration, we now reconstruct the final gradient update. The logic is simple: we preserve the safe component entirely since it violates no constraints, but we only retain a fraction $k$ of the conflicting component. Mathematically, the final gradient is composed as $\mathbf{g}_{final} = \boldsymbol{\mu}_{pla}^{\perp} + k \cdot \boldsymbol{\mu}_{pla}^{\parallel}$. By using the geometric definition $\boldsymbol{\mu}_{pla}^{\perp} = \boldsymbol{\mu}_{pla} - \boldsymbol{\mu}_{pla}^{\parallel}$, we can rewrite this equation to show that the final gradient is simply the original gradient minus a correction term:
\begin{equation}
\resizebox{0.89\linewidth}{!}{$
\mathbf{g}_{final} = (\boldsymbol{\mu}_{pla} - \boldsymbol{\mu}_{pla}^{\parallel}) + k \cdot \boldsymbol{\mu}_{pla}^{\parallel} = \boldsymbol{\mu}_{pla} - (1 - k) \boldsymbol{\mu}_{pla}^{\parallel}.
$}
\end{equation}
To make this physically intuitive, we define a new term $\alpha = 1 - k$, which represents the projection strength (i.e., how much of the conflict we strictly remove). Substituting the precision terms derived earlier, $\alpha$ takes the form $\alpha = \lambda_{sta} / (\lambda_{pla} + \lambda_{sta})$. Finally, by plugging in the formula for the conflicting component, we arrive at the final update:
\begin{equation}
\boxed{ \mathbf{g}_{final} = \boldsymbol{\mu}_{pla} - \alpha \frac{\boldsymbol{\mu}_{pla} \cdot \boldsymbol{\mu}_{sta}}{\|\boldsymbol{\mu}_{sta}\|^2} \boldsymbol{\mu}_{sta} }
\end{equation}
This closed-form solution reveals that PCR is essentially a soft projection algorithm where the projection strength $\alpha$ is not manually set but automatically calculated from the data. We can validate its rationality by examining two extreme cases. First, consider the case of a high certainty constraint where the stability gradient is very precise and rigid ($\lambda_{sta} \gg \lambda_{pla}$). Here, $\alpha$ approaches 1, meaning PCR converges to PCGrad and performs a hard projection to strictly eliminate the conflict. Conversely, consider the case of a low certainty constraint where the stability signal is noisy and unreliable ($\lambda_{sta} \ll \lambda_{pla}$). Here, $\alpha$ approaches 0, meaning PCR converges to standard gradient addition, effectively ignoring the constraint to allow full exploration. In this way, PCR achieves a smooth and theoretically grounded interpolation between obeying the constraint and ignoring it, based purely on which signal is more trustworthy.

\subsection{Policy Optimization}

While PCR provides a theoretically optimal solution for resolving conflicts, applying it element-wise to every single parameter in a billion-parameter LLM is computationally too expensive. Maintaining variance estimators and calculating projection coefficients for the entire model would drastically increase memory usage and slow down training. To make PCR practical for large-scale models, we introduce an efficient hybrid update strategy grounded in the specific roles of Transformer components. 

Modern LLMs are built from alternating Self-Attention layers and Feed-Forward Networks (MLPs). Recent interpretability studies~\cite{geva2020transformer, meng2022locating} offer a crucial insight: MLP layers act as factual knowledge stores, holding the vast majority of the model's domain expertise. In contrast, Attention layers serve primarily as context routers that move information around. Consequently, the problem of catastrophic forgetting, which represents the loss of model stability, is possibly caused by overwriting the knowledge stored within these MLP layers.

Based on this insight, we apply a strategic split. We use the computationally intensive PCR exclusively for the MLP layers to rigorously protect knowledge where it actually lives. For the Attention layers and other parameters (like LayerNorm), we revert to the standard, efficient GRPO update (simple addition). This approach strikes an optimal balance: it secures the core knowledge modules without paying the computational price for the whole model. Mathematically, the update rule for parameters $\theta^{(l)}$ at layer $l$ is defined as:
\begin{equation}
\mathbf{g}_{update}^{(l)} = 
\begin{cases} 
\mathbf{g}_{final}^{(l)} \; (\text{via Eq. 17}) & \text{if } \theta^{(l)} \in \text{MLP} \\
\mathbf{g}_{pla}^{(l)} + \beta \mathbf{g}_{sta}^{(l)} & \text{otherwise}
\end{cases}
\end{equation}
The parameters are then updated using the optimizer: $\theta^{(l)}_{t+1} \leftarrow \theta^{(l)}_{t} - \eta \cdot \text{Optimizer}(\mathbf{g}_{update}^{(l)})$. By restricting the overhead of PCR to only the most critical layers, we ensure training stability with negligible cost. The complete algorithm is detailed in Appendix D.

\section{Theoretical Analysis}
\label{sec:theory}

In this section, we demonstrate that the projection coefficient derived in PCR is not a heuristic design but the precise analytical solution under the Minimum Mean Square Error (MMSE) criterion. This means that PCR mathematically finds the optimal balance between the bias caused by violating constraints and the variance caused by noisy gradients.

\begin{theorem}[MMSE Optimality of Soft Projection]
\label{thm:optimality}
Consider the scalar estimation problem along the conflict axis defined by the unit vector $\mathbf{u} = -\boldsymbol{\mu}_{sta}/\|\boldsymbol{\mu}_{sta}\|$. Let the latent true update $z^* \in \mathbb{R}$ follow a prior distribution governed by the stability constraint $z^* \sim \mathcal{N}(0, \sigma_{sta}^2)$. Let the plasticity gradient provide a noisy observation $z_{obs} = z^* + \epsilon$, where the noise $\epsilon \sim \mathcal{N}(0, \sigma_{pla}^2)$ is independent of $z^*$.
For the family of linear estimators $\hat{z}(\alpha) = (1-\alpha)z_{obs}$ with $\alpha \in [0,1]$, the projection coefficient $\alpha^* = \frac{\lambda_{sta}}{\lambda_{pla} + \lambda_{sta}}$ utilized by PCR achieves the global minimum of the posterior expected risk $\mathcal{R}(\alpha) = \mathbb{E}[(\hat{z}(\alpha) - z^*)^2]$.
\end{theorem}

The proof is provided in Appendix G. Based on this theorem, we verify why PCR is superior to fixed strategies. By substituting $\alpha^*$ back into the risk function, the minimum risk of PCR is $\mathcal{R}_{PCR} = (\lambda_{pla} + \lambda_{sta})^{-1}$. In contrast, PCGrad (Hard Projection, $\alpha=1$) yields a risk of $\sigma_{sta}^2$, while Naive Sum (No Projection, $\alpha=0$) yields a risk of $\sigma_{pla}^2$. Since the combined precision is always higher than individual precisions, PCR is theoretically guaranteed to achieve lower error than both baselines, i.e., $\mathcal{R}_{PCR} < \mathcal{R}_{PCGrad}$ and $\mathcal{R}_{PCR} < \mathcal{R}_{Sum}$. The mechanism is physically intuitive. If the stability constraint is unreliable (large $\sigma_{sta}^2$), hard projection introduces a massive error known as bias. PCR avoids this by automatically reducing $\alpha$. Conversely, if the plasticity gradient is unstable (large $\sigma_{pla}^2$), a direct update introduces a massive error known as variance. PCR suppresses this noise by automatically increasing $\alpha$. Therefore, we can conclude that PCR guarantees better convergence stability than fixed strategy.

\begin{figure}[t]
    \centering
    \includegraphics[width=0.85\linewidth]{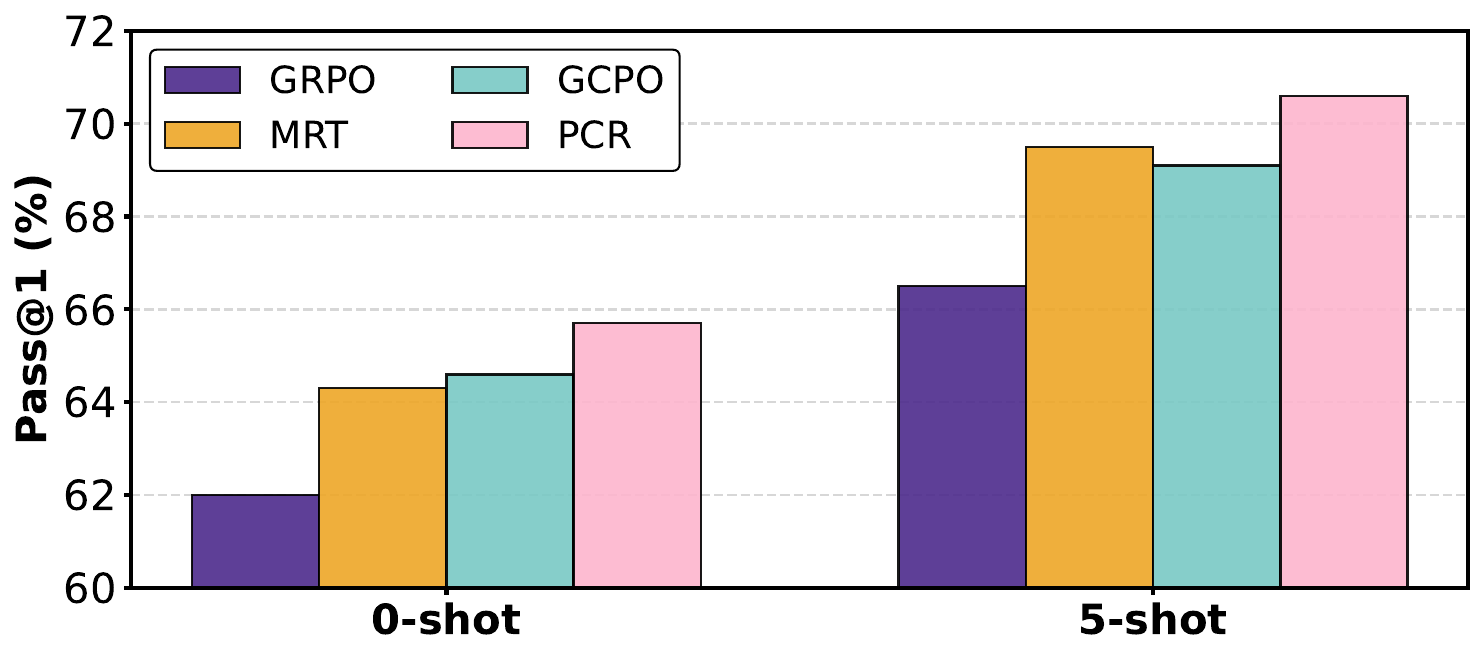}
    \caption{Performance analysis with PCR on code reasoning tasks. We record the 1-shot and 5-shot results on HumanEval.}
    \vspace{-0.15in}
    \label{fig:code}
\end{figure}

\begin{table*}[t]
  \centering
  \small
  \caption{Pass@1 performance on various math reasoning benchmarks. We compare base models trained with different fine-tuning approaches. The best results are highlighted in \textbf{bold}. More details and results are provided in Appendix L.1.}
  \label{tab:ex_1}
  % \resizebox{\linewidth}{!}{%
    \begin{tabular}{l|c|c|c|c|c|c}
      \toprule
      Base model + Method & AIME 2024 & AIME 2025 & AMC 2023 & MATH500 & MinervaMATH & \textbf{Avg.} \\
      \midrule
      \!\textbf{DeepScaleR‑1.5B‑Preview} & 42.8 & 36.7 & 83.0 & 85.2 & 24.6 & 54.5 \\
      \!~~GRPO \cite{shao2024deepseekmath} & 44.5 (+1.7) & 39.3 (+2.6) & 81.5 (-1.5) & 84.9 (-0.3) & 24.7 (+0.1) & 55.0 (+0.5) \\
      \!~~GVPO \cite{zhang2025gvpo} & 46.1 (+3.3) & 39.7 (+3.0) & 83.6 (+0.6) & 85.7 (+0.5) & 25.3 (+0.7) & 56.1 (+1.6) \\
      \!~~MRT \cite{qu2025optimizing} & 47.2 (+4.4) & 39.7 (+3.0) & 83.1 (+0.1) & 85.1 (-0.1) & 24.2 (-0.4) & 55.9 (+1.4) \\
       \!~~GCPO \cite{gu2025group} & 46.7 (+3.9) & 40.3 (+3.6) & 84.1 (+1.1) & 86.3 (+1.1) & 25.9 (+1.4) & 56.8 (+2.3) \\
      \rowcolor{blue!10} \!~~PCR (Ours) & \textbf{48.1 (+5.3)} & \textbf{41.4 (+4.7)} & \textbf{84.9 (+1.9)} & \textbf{87.0 (+1.8)} & \textbf{26.8 (+2.2)} & \textbf{57.7 (+3.2)} \\
      
      \midrule 
      \!\textbf{DeepSeek-R1-Distill-Qwen-1.5B} & 28.7 & 26.0 & 69.9 & 80.1 & 19.8 & 44.9 \\
      \!~~GRPO \cite{shao2024deepseekmath} & 29.8 (+1.1) & 27.3 (+1.3) & 70.5 (+0.6) & 80.3 (+0.2) & 22.1 (+2.3) & 46.0 (+1.1) \\
      \!~~GVPO \cite{zhang2025gvpo} & 30.6 (+1.9) & 28.2 (+2.2) & 71.5 (+1.6) & 80.5 (+0.4) & 23.1 (+3.3) & 46.7 (+1.8) \\
      \!~~MRT \cite{qu2025optimizing} & 30.3 (+1.6) & 29.3 (+3.3) & 72.9 (+3.0) & 80.4 (+0.3) & 22.5 (+2.7) & 47.1 (+2.2) \\
      \!~~GCPO \cite{gu2025group} & 31.0 (+2.3) & 29.0 (+3.0) & 71.8 (+1.9) & 81.6 (+1.5) & 23.4 (+3.6) & 47.4 (+2.5) \\
      \rowcolor{blue!10} \!~~PCR (Ours) & \textbf{32.3 (+3.6)} & \textbf{30.4 (+4.4)} & \textbf{72.7 (+2.8)} & \textbf{82.2 (+2.1)} & \textbf{24.8 (+5.0)} & \textbf{48.5 (+3.6)} \\

         \midrule
    \!\textbf{DeepSeek-R1-Distill-Qwen-7B} & 55.5 & 50.2 & 85.1 & 87.4 & 42.1 & 64.1 \\
    \!~~GRPO \cite{shao2024deepseekmath} & 56.9 (+1.4) & 51.7 (+1.5) & 85.5 (+0.4) & 87.7 (+0.3) & 43.5 (+1.4) & 65.1 (+1.0) \\
    \!~~GVPO \cite{zhang2025gvpo}       & 57.5 (+2.0) & 52.1 (+1.9) & 86.3 (+1.2) & 88.5 (+1.1) & 44.2 (+2.1) & 65.7 (+1.6) \\
    \!~~MRT \cite{qu2025optimizing} & 57.0 (+1.5) & 52.4 (+2.2) & 86.0 (+0.9) & 88.4 (+1.0) & 44.3 (+2.2) & 65.6 (+1.5) \\
    \!~~GCPO \cite{gu2025group} & 58.3 (+2.8) & 53.0 (+2.8) & 87.3 (+2.2) & 89.1 (+1.7) & 45.0 (+2.9) & 66.5 (+2.4) \\
      \rowcolor{blue!10} \!~~PCR (Ours) & \textbf{59.7 (+4.2)} & \textbf{54.1 (+3.9)} & \textbf{88.0 (+2.9)} & \textbf{89.8 (+2.4)} & \textbf{46.5 (+4.4)} & \textbf{67.6 (+3.5)} \\
      \bottomrule
    \end{tabular}%
  % }
\end{table*}

\section{Experiment}
\label{sec:experiment}
In this section, we begin by describing the experimental setup. We then present the results of the evaluation, followed by an ablation study to analyze how it works well. More details and results are provided in the Appendices J-L.

\subsection{Experimental Settings}
We evaluate our method on a diverse suite of reasoning benchmarks, covering competition math and code generation, including AIME24–25, AMC, MATH500 \cite{hendrycks2021measuring}, MinervaMATH \cite{lewkowycz2022solving}, and HumanEval \cite{chen2021evaluating}. The experiments are conducted with four base models: DeepScaleR-1.5B-Preview, DeepSeek-R1-Distill-Qwen-1.5B, DeepSeek-R1-Distill-Qwen-7B, and Qwen2-7B-Instruct. We compare against representative classic and state-of-the-art (SOTA) RL post-training baselines, including GRPO \cite{shao2024deepseekmath}, GVPO \cite{zhang2025gvpo}, MRT \cite{qu2025optimizing}, and GCPO \cite{gu2025group}.
For data preparation, DeepScaleR-1.5B-Preview, which was previously fine-tuned on 40k math QA pairs, is further trained on 919 AIME problems spanning 1989–2023. DeepSeek-R1-Distill-Qwen-1.5B is fine-tuned on a random subset of 4,000 QA pairs from NuminaMath \cite{li2024numinamath}. Following \cite{wang2025learning}, we cap both training and evaluation with a token budget of 16,384. Unless otherwise noted, we use a learning rate of $1\mathrm{e}{-6}$, weight decay of 0.01, and batch size of 256. All experiments are run on A100 GPU clusters.

\subsection{Performance Analysis}
We evaluate our method, PCR, against all the mentioned baselines across diverse benchmarks. The evaluation involves three widely adopted base models, including DeepScaleR-1.5B-Preview, DeepSeek-R1-Distill-Qwen-1.5B, and DeepSeek-R1-Distill-Qwen-7B, with performance measured by pass@1 accuracy following the protocols in \cite{gu2025group}. 
The results are shown in Table \ref{tab:ex_1}.
Across all experimental settings, our method consistently secures the top performance, surpassing both the base models and competitive RL baselines. Specifically, our method yields substantial improvements over the base models, achieving average gains of 3\% on all the base models.
Critical to this success is our Bayesian formulation, which effectively integrates generalizability and transferability. Unlike standard RL baselines that may overfit to specific reward signals, e.g., GRPO with negative performance gains in AMC, our method treats the optimization process as a Bayesian inference problem, i.e., our method demonstrates superior robustness on challenging benchmarks such as AIME and MinervaMATH, where it outperforms the strongest baselines by over nearly 1.2\%.
These results demonstrate the effectiveness of our method.

To further validate the versatility of PCR, we conduct experiments on code reasoning tasks using Qwen2-7B-Instruct. We evaluate performance via the standard HumanEval protocol under both 0-shot and 5-shot settings. The results are presented in Figure~\ref{fig:code}.
From the results, we can observe that our method achieves the best results among all methods: PCR achieves the improvements of almost 1.2\% over the baselines. 
This further shows the advantages of our method.

\begin{figure}[t]
    \centering
    \includegraphics[width=0.82\linewidth]{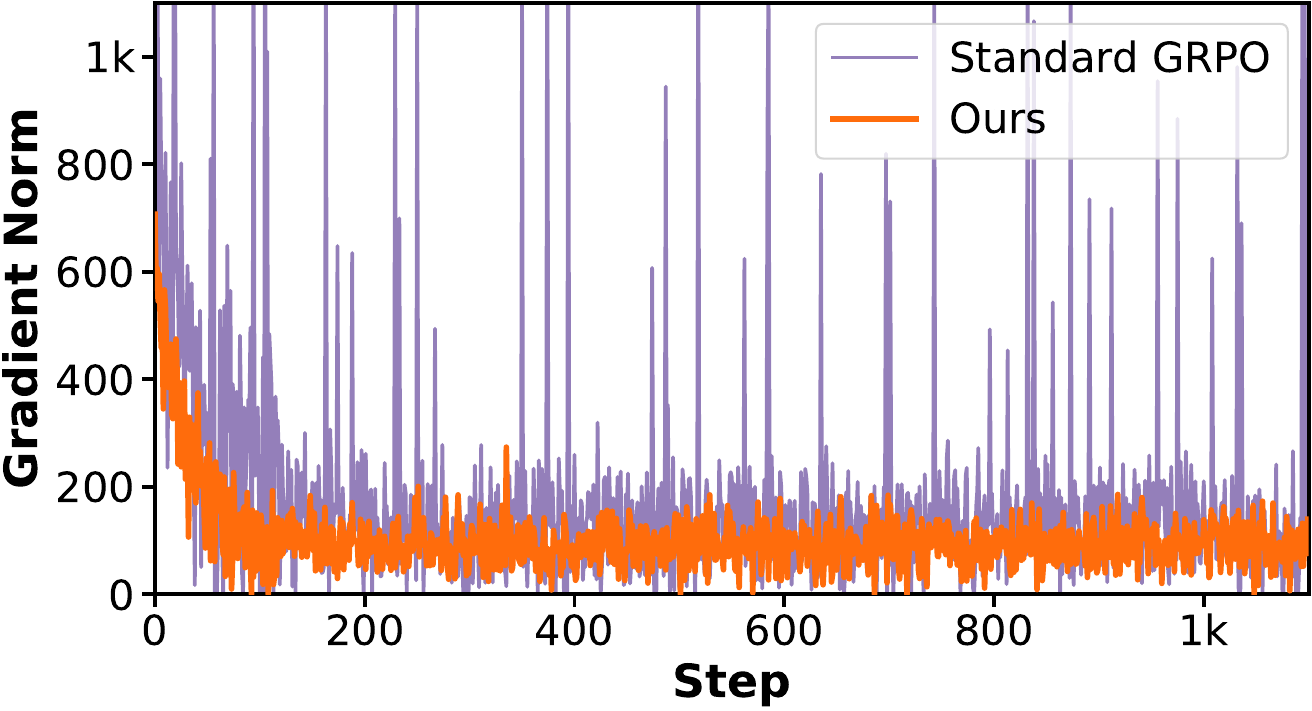}
    \caption{Stability analyses. We provide the norm of the gradient during training. A stable gradient norm implies consistent updates; large swings suggest unstable or overly aggressive shifts.}
    \vspace{-0.15in}
    \label{fig:stablity}
\end{figure}

\begin{figure*}
    \centering

    \begin{subfigure}[b]{0.33\textwidth}
    \centering
    \includegraphics[width=0.9\textwidth]{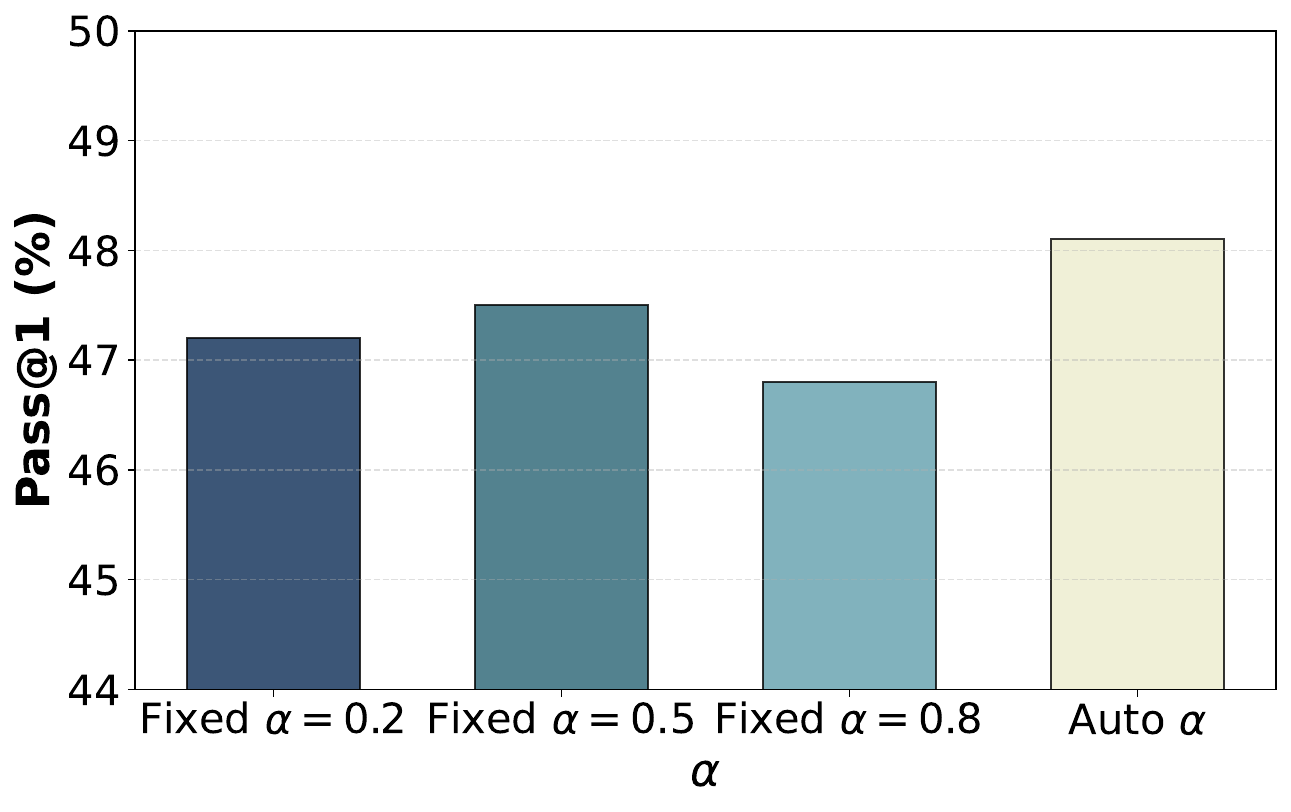}
    % \label{fig:gradient_roles}
    \caption{Effect of probabilistic modeling.}
  \end{subfigure}
  \hfill
  \begin{subfigure}[b]{0.33\textwidth}
    \centering
    \includegraphics[width=0.9\textwidth]{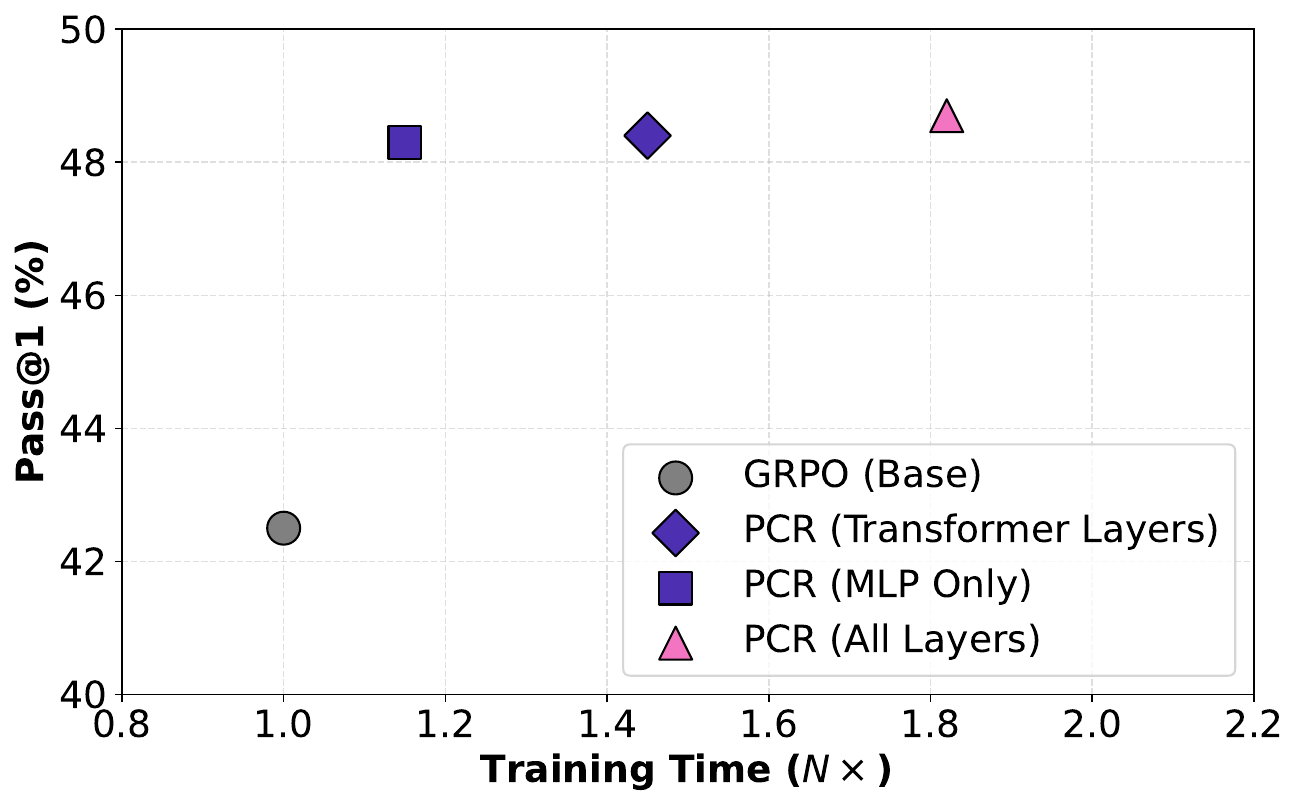}
    \caption{Effect of hybrid strategy.}
  \end{subfigure}
  \hfill 
  \begin{subfigure}[b]{0.33\textwidth}
    \centering
    \includegraphics[width=0.9\textwidth]{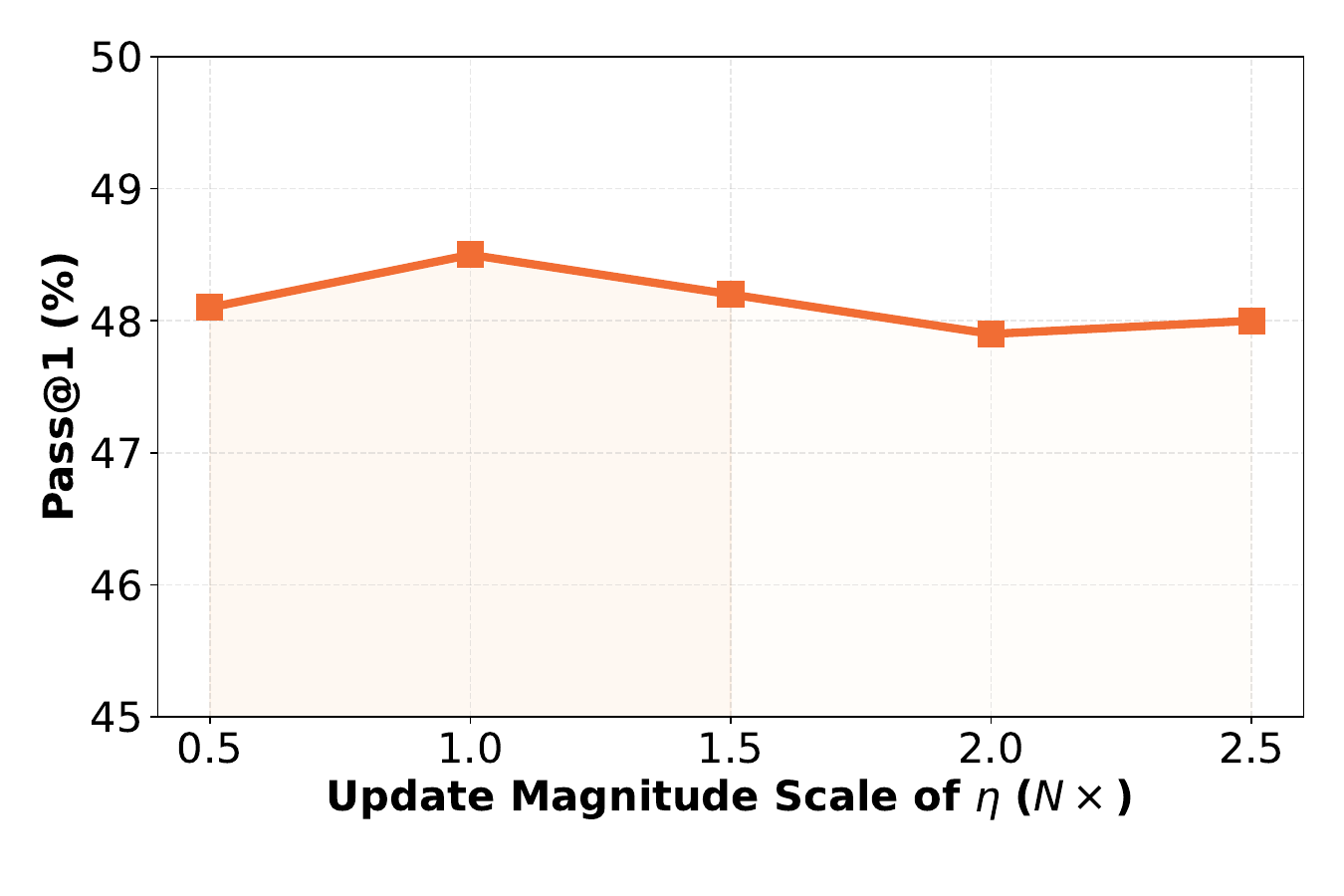}
    \caption{Impact of Learning Rate $\eta$.}
    \label{fig:scaling}
  \end{subfigure}

    \caption{Ablation Study. (a) and (b) evaluate the effect of different components within PCR. (c) shows the scalability to larger update magnitudes. More experiments and results are provided in Appendix L.}
    \label{fig:ablation}
    \vspace{-0.1in}
\end{figure*}

\begin{figure}[t]
    \centering
    \includegraphics[width=\linewidth]{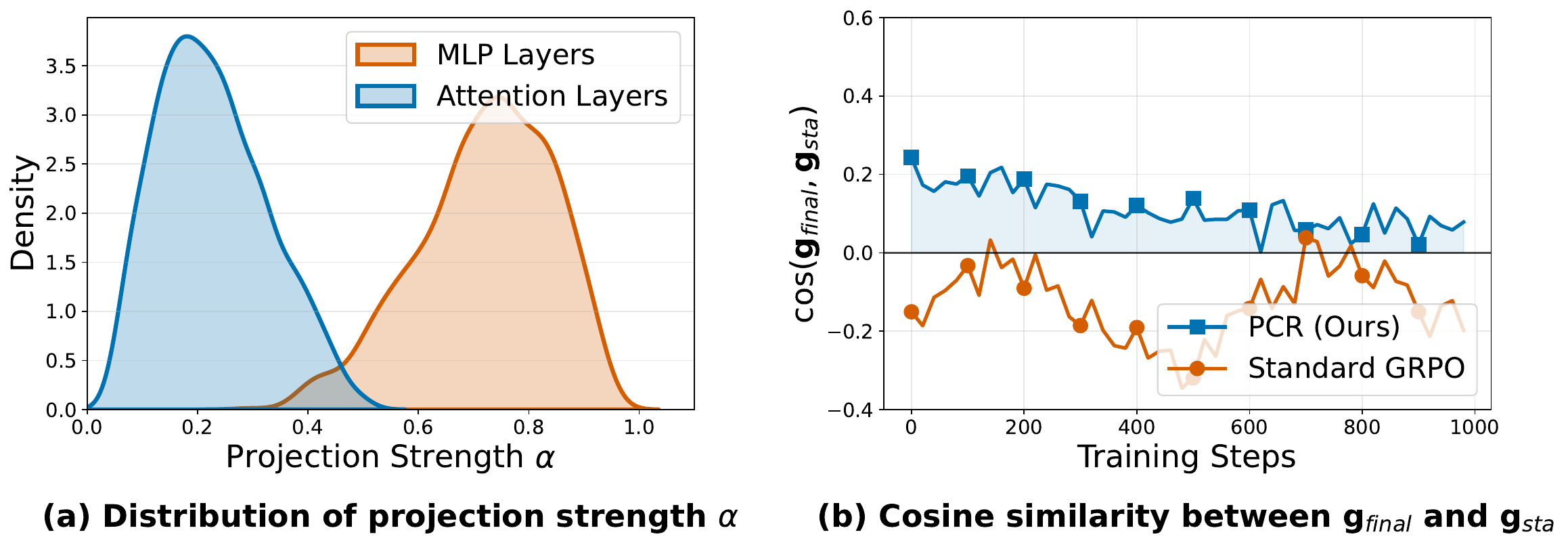}
    \caption{Visualization results.
    (a) The distribution of projection strength. (b) The cosine similarity between $\mathbf{g}_{final}$ and $\mathbf{g}_{sta}$. 
    More results are shown in Appendix L.2.}
    \vspace{-0.1in}
    \label{fig:vis}
\end{figure}

\subsection{Visualization Analysis}
To further investigate how PCR dynamically resolves gradient conflicts, we conduct a series of visualization experiments. 
We first visualize the statistical distribution of the projection strength $\alpha$ across different functional modules during training (Figure \ref{fig:vis}(a)). Unlike PCGrad, which applies a binary projection (i.e., $\alpha \in \{0, 1\}$), PCR exhibits a continuous and adaptive distribution. Notably, we observe that MLP layers generally maintain a higher $\alpha$ density compared to Attention layers. This confirms our hypothesis that knowledge-heavy MLP modules encounter more severe directional conflicts, requiring stronger Bayesian arbitration to safeguard linguistic stability.
We then measure the cosine similarity between the final update gradient $\mathbf{g}_{final}$ and the stability gradient $\mathbf{g}_{sta}$ throughout the post-training phase. 
As shown in Figure \ref{fig:vis}(b), PCR maintains a non-negative or slightly positive correlation.
Besides, we also follow the same experimental settings in Subsection \ref{sec:empirical_findings}, and record the corresponding results. The results in Figure 6 shows that our method breaks the suboptimal Pareto frontier constrained by standard GRPO: PCR achieves superior AIME accuracy improvement while simultaneously maintaining WikiText-2 PPL. These results demonstrate that PCR successfully navigates the underlying conflicts, improving the LLM performance. 
See Appendix L.2 for more results.

\subsection{Training Stability Analyses}
% fig:stablity
Given the high cost of post-training LLMs, optimization stability is essential for efficient convergence and avoiding collapse. Instability often stems from gradient conflict between reward maximization and reference preservation, which amplifies update variance and wastes compute. To quantitatively assess this, we utilize the gradient norm as a proxy for optimization smoothness, consistent with standard RL practices \cite{xiao2025bnpo}.
As shown in Figure \ref{fig:stablity}, PCR exhibits the most stable training dynamics among all compared methods. While GRPO suffers from pronounced oscillations due to the conflicting gradients, PCR maintains a relatively smooth and consistent gradient norm throughout the training process. This superior stability demonstrates the effectiveness of our Bayesian fusion mechanism.

\subsection{Ablation Studies}

\textbf{The effect of different components within PCR.} 
We conduct ablation studies to justify the necessity of each component in PCR. 
Specifically, we first compare PCR (i.e., auto $\alpha$) against fixed soft projection baselines (i.e., $\alpha \in \{0.2, 0.5, 0.8\}$). The results in Figure \ref{fig:ablation}(a) demonstrate that our method consistently outperforms any fixed heuristic, proving the superiority of Bayesian arbitration. Next, we evaluate applying PCR to different layer subsets. Our results in Figure \ref{fig:ablation}(b) show that applying PCR exclusively to MLP layers achieves a performance gain comparable to the all layers setting while significantly reducing training time. This demonstrate the effectiveness of our design.

\textbf{Parameter Sensitivity.}
We also conduct sensitivity studies on the weighting coefficient $\beta$ and the PCR parameter learning rate $\eta$. Figures 6a and~\ref{fig:scaling} show that PCR is highly robust to variations in $\beta$ and continues to improve performance as the parameter update magnitude increases.

\section{Conclusion}
\label{sec:conclusion}

This work attributes the instability of GRPO to the high-dimensional conflict between task-specific optimization and reference maintenance. We argue that standard gradient aggregation fails due to the stochastic nature of group sampling. To resolve this, we introduced Probabilistic Conflict Resolution (PCR), which reformulates gradient projection as a Bayesian inference problem. By adaptively balancing exploration bias against constraint variance, PCR acts as an optimal linear filter in the gradient space. Our results confirm the effectiveness of the proposed PCR.

\section*{Impact Statement}
This paper presents work whose goal is to advance the field of machine learning. There are many potential societal consequences of our work, none of which we feel must be specifically highlighted here.

\bibliography{main}
\bibliographystyle{icml2026}

\end{document}